\documentclass[letterpaper, 10 pt, conference]{ieeeconf}  % Comment this line out if you need a4paper

\IEEEoverridecommandlockouts                              % This command is only needed if
\usepackage{algorithm}
\usepackage{algpseudocode}
\usepackage{graphicx}
\usepackage[utf8]{inputenc} % allow utf-8 input
\usepackage[T1]{fontenc}    % use 8-bit T1 fonts
\usepackage{url}            % simple URL typesetting
\usepackage{booktabs}       % professional-quality tables
\usepackage{amsfonts}       % blackboard math symbols
\usepackage{nicefrac}       % compact symbols for 1/2, etc.
\usepackage{microtype}      % microtypography
\usepackage{color}
\usepackage{subfigure}
\usepackage{caption}
\usepackage{tikz}
\usepackage{paralist}

\usepackage{booktabs} % for professional tables
\usepackage{placeins}
\usepackage{color}

\usepackage{amsmath}
\usepackage{algorithm}
\usepackage{algpseudocode}
\usepackage{array,booktabs}
\usepackage{flushend}

\usepackage{multirow}

\usepackage{tabularx}
\usepackage{adjustbox}
\usepackage{rotating}

\usepackage{afterpage}
\usepackage{longtable}
\usepackage{booktabs}
\usepackage{arydshln}

\makeatletter
\let\NAT@parse\undefined
\makeatother

\usepackage{xcolor}
\definecolor{webred}{rgb}{0.5,0,0}
\definecolor{webblue}{rgb}{0,0,0.8}
\usepackage[pdftex,colorlinks,citecolor=webblue,linkcolor=webred,]{hyperref}

\title{\LARGE \bf
Stress Testing Autonomous Racing Overtake Maneuvers with RRT
}

\author{Stanley Bak$^{\dagger*}$, Johannes Betz$^{\ddagger*}$, Abhinav Chawla$^{\dagger*}$, Hongrui Zheng$^{\ddagger*}$, and Rahul Mangharam$^{\ddagger}$
\thanks{$^{\dagger}$Stony Brook University, Department of Computer Science,
        11794 Stony Brook, NY, USA
        {\tt\small stanley.bak, abhinav.chawla}@stonybrook.edu}
\thanks{$^{\ddagger}$University of Pennsylvania, School of Engineering and Applied Science,
        19106 Philadelphia, PA, USA        
        {\tt\small joebetz, hongruiz, rahulm}@seas.upenn.edu}%
\thanks{$^*$Authors contributed equally.}
}

\begin{document}

\newtheorem{definition}{Definition}

\maketitle
\thispagestyle{empty}
\pagestyle{empty}

\begin{abstract}
High-performance autonomy often must operate at the boundaries of safety.
When external agents are present in a system, the process of ensuring safety without sacrificing performance becomes extremely difficult.
In this paper we present an approach to stress test such systems based on the rapidly exploring random tree (RRT) algorithm.

We propose to find faults in such systems through \emph{adversarial agent perturbations}, where the behaviors of other agents in an otherwise fixed scenario are modified.
This creates a large search space of possibilities, which we explore both randomly and with a focused strategy that runs RRT in a bounded projection of the observable states that we call the \emph{objective space}.
The approach is applied to generate tests for evaluating overtaking logic and path planning algorithms in autonomous racing, where the vehicles are driving at high speed in an adversarial environment.
We evaluate several autonomous racing path planners, finding numerous collisions during overtake maneuvers in all planners.
The focused RRT search finds several times more crashes than the random strategy, and, for certain planners, tens to hundreds of times more crashes in the second half of the track.
%
%Based on this evaluation we can afterwards present a holistic summary about the flaws in the path planning and control approach.
\end{abstract}

% why not mention disparity extender?

%, and  variety of faults, and demonstrate
%where the vehicles are driving on high speeds in an adversarial environment. 
%and apply the approach to overtake maneuvers in autonomous racing.
%
%Our approach is based on a modification to 
%While natural tests, as well as scenario-based testing may occasionally result in overtake

%Rather than testing the system on multiple distinct scenarios, as is often done with autonomous driving, we instead develop methods to explore the system's reaction to subtle changes in the opponent vehicles for one specific scenario.

%Autonomous vehicles will operate completely without a safety driver in the near future.
%
%This means the cars have to make decisions under uncertainty in difficult situations were a crash can happen fast and easily.
%
%The underlying path planner therefore needs to provide a fast and reliable trajectory that is maneuvering the vehicle through complicated scenarios. 
%
%We therefore present a pipeline for testing path planning algorithms for autonomous vehicles.
%
\section{Introduction}

% indy autonomous challenge
% $1.5 million prize competition among universities
% announced october 2019 -> started officialy march 2020
% June 30, 2021 Ansys
% competition: 23rd october
% citations: https://linklab-uva.github.io/icra-autonomous-racing/contributed_papers/paper1.pdf
% https://linklab-uva.github.io/icra-autonomous-racing/contributed_papers/paper6.pdf

\begin{figure}[t]
    \includegraphics[scale=0.9]{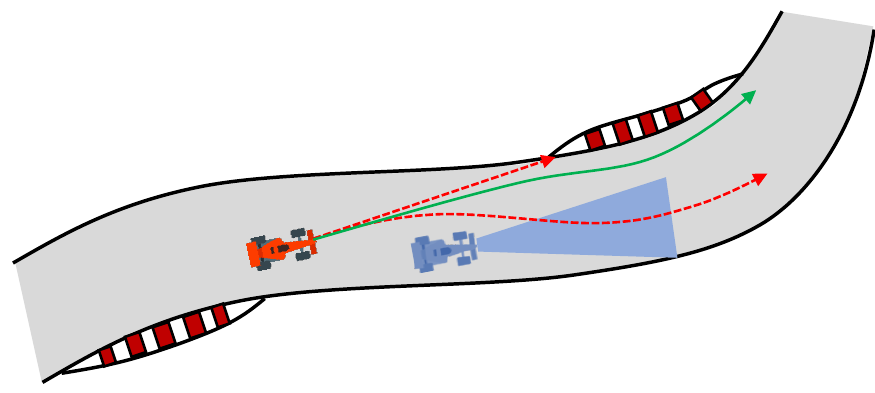}
    \caption{Illustrative overview of autonomous path planning on the racetrack for two vehicles.}
    \label{fig:racing}
    \vspace{-20pt}
\end{figure}

Controlling autonomous systems near their performance boundaries and in the presence of adversarial agents is hard.
Consider the Indy Autonomous Challenge~\cite{indy}, a full-size autonomous racing competition with over 30 university teams originally announced in 2019 and scheduled to take place in October 2021.
In this environment, vehicles are driving at high speeds with high accelerations in an adversarial environment. 
Overtaking maneuvers like the one shown in Figure \ref{fig:racing} are difficult to plan because the safety of the maneuver depends on the future actions of the opponent, the interaction between the vehicles, different topological constraints at each track segment and the correct observance of the vehicle dynamic limits.
Even with the large incentives to create safe and high performance autonomy---including a \$1.5 million prize pool---there are strong reasons to be suspicious of the safety of the autonomous logic during overtake.
In the most recent simulated race in June 2021\footnote{\url{https://www.youtube.com/watch?v=gTjQ3sWdYh0}}~\cite{ansys}, in the single-car qualification stage, most vehicles performed extremely well---the lap time for the top 14 of the 16 cars was within half a second.
However, during multi-car races, 12 of the 16 competing teams were disqualified for causing crashes with other vehicles or the wall.
%
%The winning team had the fastest time during qualification and could start at the front, and all its races were won without demonstrating an overtake maneuver.

Part of the reason for the difficulty of creating safe overtake logic is that traditional end-to-end scenario testing is inefficient.
The number of situations where the logic must perform well is vast, but only a small number of interactions are exposed during each long simulation run of a single scenario.
In this paper, we strive to address this shortcoming by developing targeted test-generation approaches.
Rather than testing a large number of scenarios (starting car positions, different racetracks, number of opponents), we instead explore the large number of possibilities within a single scenario. 
We focus on detecting a wide variety of crashes during autonomous overtaking maneuvers.

To do this, we perform \emph{adversarial agent perturbations}, slightly modifying the opponent vehicle behaviors at different points in time.
For example, an opponent vehicle can be altered to drive slightly faster or slightly slower.
Repeatedly performing small perturbations creates a large tree of possible simulations, which we then explore to look for collisions.
To do this without duplicating simulation effort, our search strategy is based on rapidly exploring random trees (RRT)~\cite{Lavalle98rapidly-exploringrandom}. 
Further, we target the search on
a projection of the full simulation state space where overtake maneuvers are likely to occur, which we call the \emph{objective space}.

We evaluate our approach using quantitative autonomy coverage metrics based on spatial clustering approaches.
For five different planning algorithms, we show our approach has superior performance compared with random exploration of the adversarial perturbation search tree.

The main contributions of the paper are:
\begin{itemize}
    \item The use of adversarial agent perturbations to create a tree of possible behaviors to explore;
    \item The efficient searching through the possible behaviors using a random and more efficient RRT-based strategy;
    \item The evaluation of the performance of the autonomy through quantitative failure coverage metrics;
    \item The application of our framework to five planners in the autonomous racing and overtake scenario.
\end{itemize}

The rest of this paper is organized as follows.
First we review and compare our approach with related test-generation methods in Section~\ref{sec:related}.
Next, in Section~\ref{sec:method}, we present our methodology as a generic test-generation algorithm that could be used for any autonomous system with adversaries.
In Section~\ref{sec:exp},
we specialize our generic approach to the concrete problem of overtaking in autonomous racing, using an open-source simulator~\cite{okelly2019} for the F1TENTH autonomous racing competition\footnote{\url{https://f1tenth.org}}.
The paper finishes with a discussion in Section~\ref{sec:discussion} followed by a conclusion.

\section{Related Work}
\label{sec:related}
%\subsection{Autonomous Vehicle Testing}
%this is grouped into the testing scenario search and generation part
%A well-known evaluation framework for path planning algorithms is \textit{CommonRoad} \cite{Althoff2017} which provides benchmarks which are composed by a vehicle model, a cost function, and the definition of the scenario where the motion planner is tested. Currently CommonRoad has no particular feedback on the flaws of the planer nor does it consider path controller.
%Wheeler et al \cite{Wheeler2019} presents an automated method for identifying and clustering critical situations that can occur in autonomous systems. By using importance sampling and efficiently optimize its parameters a risk-based safety validation of the system can be done.
%In \cite{Stahl2020} graphical user interface (GUI) is presented that allows the generation of a multi-vehicle path planning scenarios in a race environment. This GUI provides the possibility of scenario-based evaluation of path planners and safety modules. Unfortunately the creation of the scenarios need to be done manually and therefore not automatic test creation is provided.
%\subsection{Sampling Based Monitoring, Testing, and Verification}
The line of work closest to our approach is sampling-based testing and verification. 
RRT has been considered for test generation before~\cite{kim2006sampling}.
A new distance function, new weighting, and a scheme to modify sampling probability distribution on the fly were introduced, reducing computation time by an order of magnitude.  
However, the method samples in the full state-space of the system, and does not consider adversarial multi-agent systems that have unknown dynamics. 
In contrast, since we search a projection of the state space we call the objective space, our method is compatible with black-box systems where only the environment and agent actions are observable. 
Nonetheless, we could consider adapting some of their proposed optimizations to improve our search, such as using gradient information to improve sample selection.

Other work uses sampling-based verification algorithms to falsify temporal logic specifications of the system~\cite{cheng2008sampling}. 
This method focuses on resolution completeness of the falsification problem while we focus on finding a large variety of failure cases.
Other falsification approaches for hybrid systems~\cite{julius2007robust,annpureddy2011s,donze2010breach} try to minimize robustness metrics for temporal logics.
Robustness metrics capture how close a simulation is to an unsafe state.
This would not work well for autonomous racing collisions, as high-performance controllers typically drive close to track boundaries where robustness would be low.
These methods also generally compute full rollouts which may be slow, and work best when the input dimension of the system is modest.
Partial rollout test generation has also been considered~\cite{zutshi2014multiple,zutshi2013trajectory}, using the notion of trajectory splicing.
For these methods, an abstraction of the system is constructed based on short simulation runs.
When problems are found, it can be difficult to concretize the abstract counterexample, to connect the trajectory pieces to create an actual error witness.

Data-driven verification procedures have also been used for test generation~\cite{quindlen2018active}, which use SVM to classify whether trajectories will be in the safe or unsafe set. 
In addition, this work uses an active learning scheme to request additional sample points to update the model. 
In comparison, our method aims to find many different failure cases by sampling short rollouts of the simulation. 
We also do not suffer from generalization issues since we do not need to create a predictive model to classify the safety of system trajectories.
%\cite{bonakdarpour2011sampling} proposes formal semantics of time-triggered sampling-based monitoring and showed optimization of sampling period using minimum auxiliary memory. \bz{this one is slightly further from what we're doing compared to the others.}

%\subsection{Reachability Analysis}
Reachability analysis has also been applied to autonomous vehicles as an attempt to compute all possible future scenarios in order to verify the system's safety \cite{althoff14_online_journal, althoff2010reachability, falcone2011predictive, althoff2007reachability, althoff2008verification, asarin2006recent,jewell2021embedded}. 
This line of work requires accurate symbolic models of the target system and sometimes other agents in the same environment. 
Although some of the work can incorporate uncertainty, the problem often becomes intractable for complex dynamics or suffers from excessive error.
For this reason, applications of reachability have generally been limited to structured environments like everyday traffic.

%\subsection{Test Generation and Falsification}
Testing scenario search and generation has also been widely applied to the testing and verification of autonomous systems~\cite{anand2013orchestrated, o2017computer, rocklage2017automated, Althoff2017, Stahl2020}. 
Some approaches employ optimization or adaptive sampling techniques to accelerate finding test cases with highest risk to the system~\cite{okelly2018scalable, norden2019efficient, mullins2017automated, tuncali2018simulation, klischat2019generating, karunakaran2020efficient, Wheeler2019}.
In comparison to this line of work, our approach operates in less structured driving conditions, and we focus on finding a large variety of failures rather than rare corner cases. 
Our approach is compatible with scenario-based testing, as we focus on exploring the possibilities within each individual scenario (intra-scenario testing).

%\subsection{White Box, Gray Box, and Black Box Testing}
Lastly, in software fuzz testing, test generation can be classified into white-box, gray-box, and black-box methods, depending on whether the inner workings of the software is known to the tester~\cite{fuzzingbook2021,godefroid2008automated,nidhra2012black,khan2012comparative}.
Our application is different, in that the planner is reactive, tightly interacting with a physical environment simulator.
In this classification, our approach is hybrid of black-box and gray-box methods---we treat both the planner logic and environment as a black box, but use their interface to drive the search process.
%\subsection{Fuzz Testing}

\section{Methodology}
\label{sec:method}

We now present the test generation problem and solution framework we propose.
In this section, we describe it as a general approach that we could use to create tests for any autonomous system with adversarial agents.
%
%The method makes use of problem-specific information that we abstractly refer to as adversarial-agent perturbations and objective space projections.
%
In the next section, we will apply this generic method to the concrete problem of overtake in autonomous racing.

\subsection{Problem Statement}
\begin{definition}
	We describe the model of the system under test as a tuple $\mathcal{S}=(\mathcal{X},\mathcal{T},\mathcal{U})$ where
	\begin{itemize}
		\item $\mathcal{X}$ is a set of observable states of the system
		\item $\mathcal{T}$ is the corresponding time steps to the states $\mathcal{X}$
		\item $\mathcal{U}$ is a finite set of external inputs (adversarial perturbations) possible at each step in $\mathcal{T}$.
	\end{itemize}
\end{definition}
\begin{definition}
	We define the testing problem as: given a tuple $(\mathcal{E}, \mathcal{S},\mathcal{T}_{[i, e]},\mathcal{V},\mathcal{F})$ where
	\begin{itemize}
		\item $\mathcal{E}$ is the selected scenario where the system operates in
		\item $\mathcal{S}$ is system under test
%		\item $\mathcal{X}_{init}$ is the starting state of the system
		\item $\mathcal{T}_{[i, e]}$ is a discrete interval of testing time steps
		\item $\mathcal{V}$ is the finite set of adversarial agent perturbations
		\item $\mathcal{F}$ is the failure specification set.
	\end{itemize}
	
	The goal of testing is to find \emph{failing tests}. 
	A failing test maps adversarial agent perturbations from each time step in $\mathcal{T}_{[i, e]}$
	to $\mathcal{V}$, such that $\exists t\in \mathcal{T}_{[i, e]}$ where $\mathcal{X}(t) \in \mathcal{F}$.
	We are interested to both maximize the number of failing tests, as we well as to explore a variety of failures (defined later using spatial clustering methods).
\end{definition}

Specifically in our case, the system under test involves path planning algorithms and controllers for autonomous race cars.
The observable states of the system $\mathcal{X}$ consist of the pose $(x, y, \theta)$ of the ego vehicle in the world, the current laser scan of the ego vehicle, and the poses of other vehicles in the world.
The unobservable states of the system consists of the decision making logic and control rules used by the motion planner, as well as any internal environmental states not exposed to the planners.
The adversarial perturbation inputs $\mathcal{U}$ are modifications to the command outputs of the motion planners according to the observable states. 
The failure specification set $\mathcal{F}$ consists of all states where the ego vehicle is in collision with either the race track or other agents.

\subsection{Stress Testing Pipeline}
\begin{figure}[h]
    \vspace{-10pt}
	\includegraphics[width=\columnwidth]{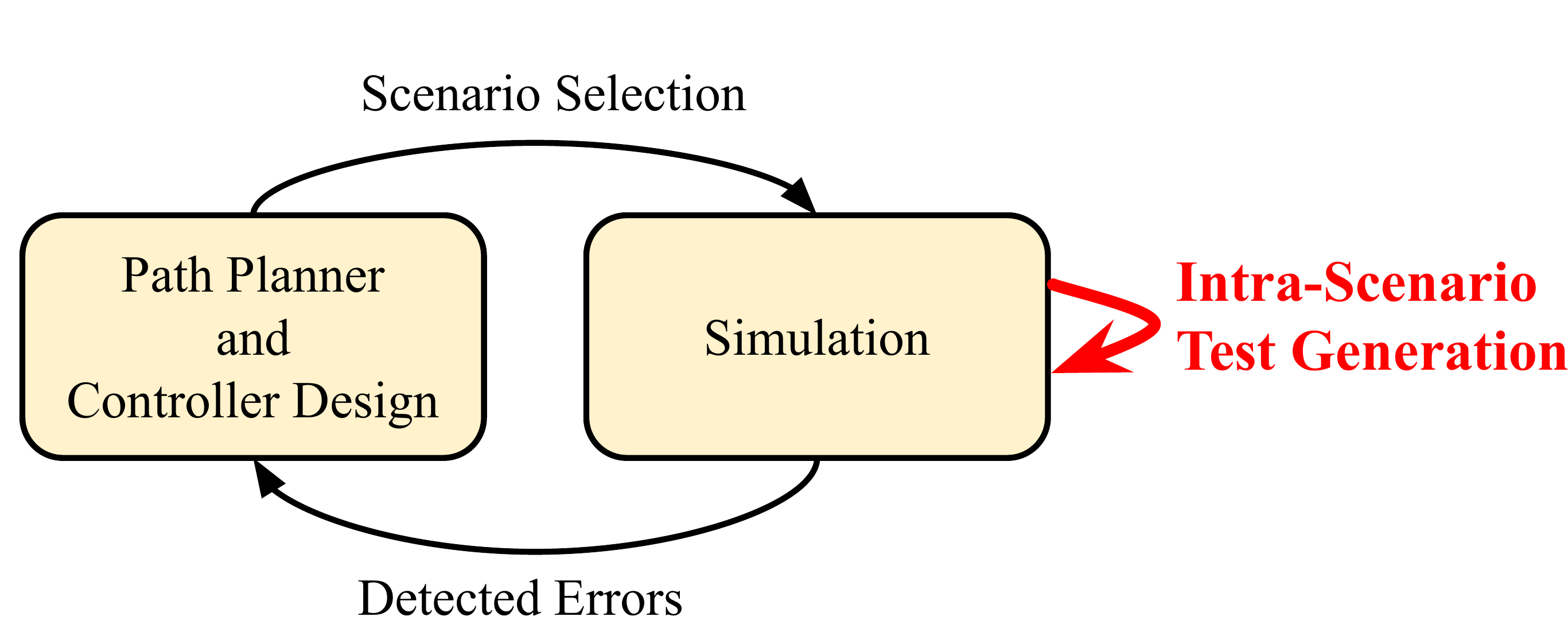}
	\caption{This work focuses on one part of the autonomous systems testing process, intra-scenario test generation (red).}
	\label{fig:pipeline}
	\vspace{-5pt}
\end{figure}

Our test generation method fits into a larger autonomy design and verification process as shown in Figure~\ref{fig:pipeline}. 

Although we do not explore scenario selection in this work, scenario selection (inter-scenario testing) and our proposed algorithm (intra-scenario testing) complement each other. 
After the development of an initial version of the path planner and controller, a test engineer (or algorithm) selects an appropriate test scenario which includes the initial states and environmental conditions. 
Our method then stress-tests this scenario by finding errors under slightly different behaviors of each of the adversarial agents (red arrow).
Lastly, the found failure cases provide feedback on the performance of the system, and can be used by the test engineer (or an algorithm) to either select new scenarios and collect further failures or to tune parameters in the path planner.

\subsection{Intra-Scenario Test Generation Details}
\label{sec:generic_stress_tester}
We now describe the details of the test generation process.
The algorithm uses the following application-specific inputs:
\begin{itemize}
    \item \textbf{Initial Scenario Configuration}: $q_{init}\in\mathbb{R}^{n}$, initial observed state to initialize the environment simulator;
    \item \textbf{Adversarial Agent Perturbations}: finite set $\mathcal{V}$ of size $m$, describes perturbations that could be made to the adversarial agents at each step;
    \item \textbf{Objective Space Projection Function}: $\mathcal{P} : \mathbb{R}^{n} \rightarrow \mathbb{R}^o$ maps observed state to the RRT search space;
    \item \textbf{Objective Space Exploration Limits}: $\mathcal{B}\in\mathbb{R}^{2o}$ describes the limits for sampling the objective space;
    \item \textbf{Simulation Step Function}:
    $\texttt{step\_sim}$, computes a short-time rollout of the simulation environment given a start state and adversarial agent perturbation.
\end{itemize}

With these, we perform RRT-based exploration to detect failing tests, shown in Algorithm~\ref{alg:cap}.
At line 1, the search is initialized by taking the start state of the scenario $q_{init}$ and projecting it to the objective space.
This serves as the root node of the tree, and the tree of explored states is initialized with the root node at line 2.
At lines 4 and 5, the algorithm then proceeds like RRT in the objective space, generating random $o$-dimensional points subject to the limits $\mathcal{B}$, finding the closest node within $\mathcal{B}$ using a normalized $L_2$-norm distance metric. And at line 6, the state corresponding to the closest node is found.
In lines 7 to 11, all adversarial perturbations $\mathcal{V}$ are then explored from this node using short rollouts performed by the \texttt{step\_sim} function, which also takes in a specific adversarial perturbation to use.
The \texttt{step\_sim} function is custom to the scenario being considered, and we elaborate on it further in the next section in the context of autonomous racing.
The resultant states are projected to the objective space using $\mathcal{P}$ and added to the search tree $G$.
We generally use stopping conditions that cap the number of explored nodes, which corresponds to the number of steps the simulation environment has executed, although a timeout could also be used.
Every node created in the process also stores all information necessary to recreate the simulation starting at the initial state, guaranteeing all states explored (and failures found) can be recreated.
The algorithm returns the $m$-ary tree of explored nodes $G$, which can then be quickly checked to provide the lists of perturbations that led to failures using the specification failure set $\mathcal{F}$.

\renewcommand{\algorithmicrequire}{\textbf{Input:}}
\renewcommand{\algorithmicensure}{\textbf{Output:}}
\begin{algorithm}[t]
\caption{RRT Based Algorithm}\label{alg:cap}
\begin{algorithmic}[31]
%\Require $\mathcal{B}, q_{init}, \mathcal{V}, \mathcal{E}_{sim}, N$
\Ensure tree of explored states $G$
%\State total\_nodes $\gets 1$
\State $p_{init} \gets \mathcal{P}(q_{init})$
\State $G$\texttt{.init}($p_{init})$
\While{not finished}
\State $p_{rand}\gets$\texttt{rand\_sample($\mathcal{B}$)}
\State $p_{nearest} \gets$\texttt{nearest}($p_{rand}$, $G$, $\mathcal{B}$)
\State $q_{nearest} \gets \mathcal{P}^{-1}(p_{nearest})$
%\State $q_{curr} \gets$\texttt{expand}($q_{nearest}$, $q_{rand}$)
\For{adv in $\mathcal{V}$ }
\State $q_{child}=$ \texttt{step\_sim}($q_{nearest}$, adv)
\State $p_{child} \gets \mathcal{P}(q_{child})$
%\State $q_{curr}$\texttt{.add\_child}($q_{child}$)
%\State $node.children[cmd]$ = $child\_node$
%\State total\_nodes$++$
\State $G$\texttt{.add\_vertex}($p_{child}$)
\State $G$\texttt{.add\_edge}($p_{nearest}$, $p_{child}$)
\EndFor

\EndWhile
\end{algorithmic}
\end{algorithm}

\begin{figure*}[t]
    \centering
    \includegraphics[clip, trim=0.5cm 3.5cm 0.1cm 3.5cm, width = 0.4\textwidth]{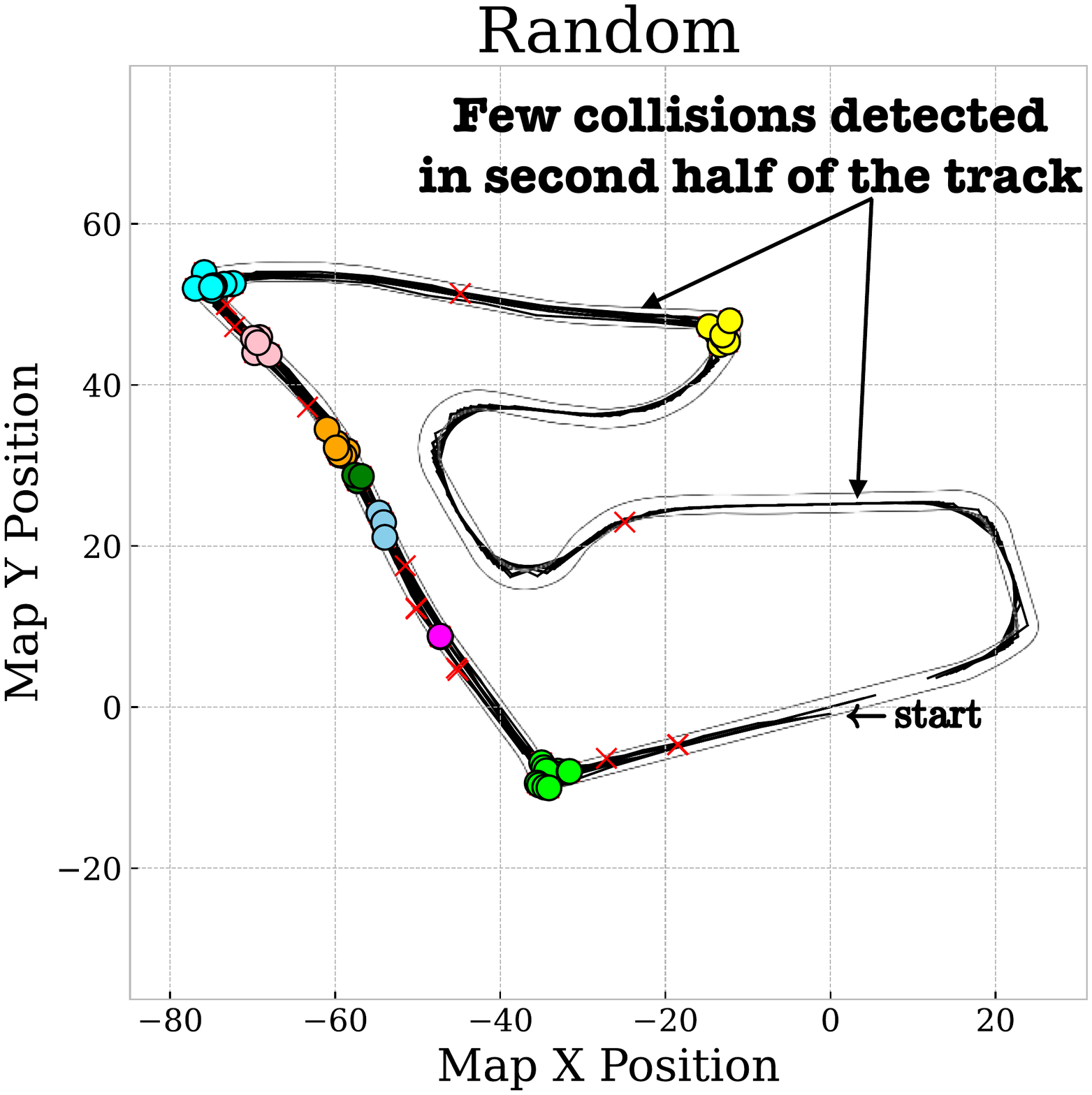}
    ~~~~~~~~~~~~~~~
    \includegraphics[clip, trim=0.5cm 3.5cm 0.1cm 3.5cm, width = 0.4\textwidth]{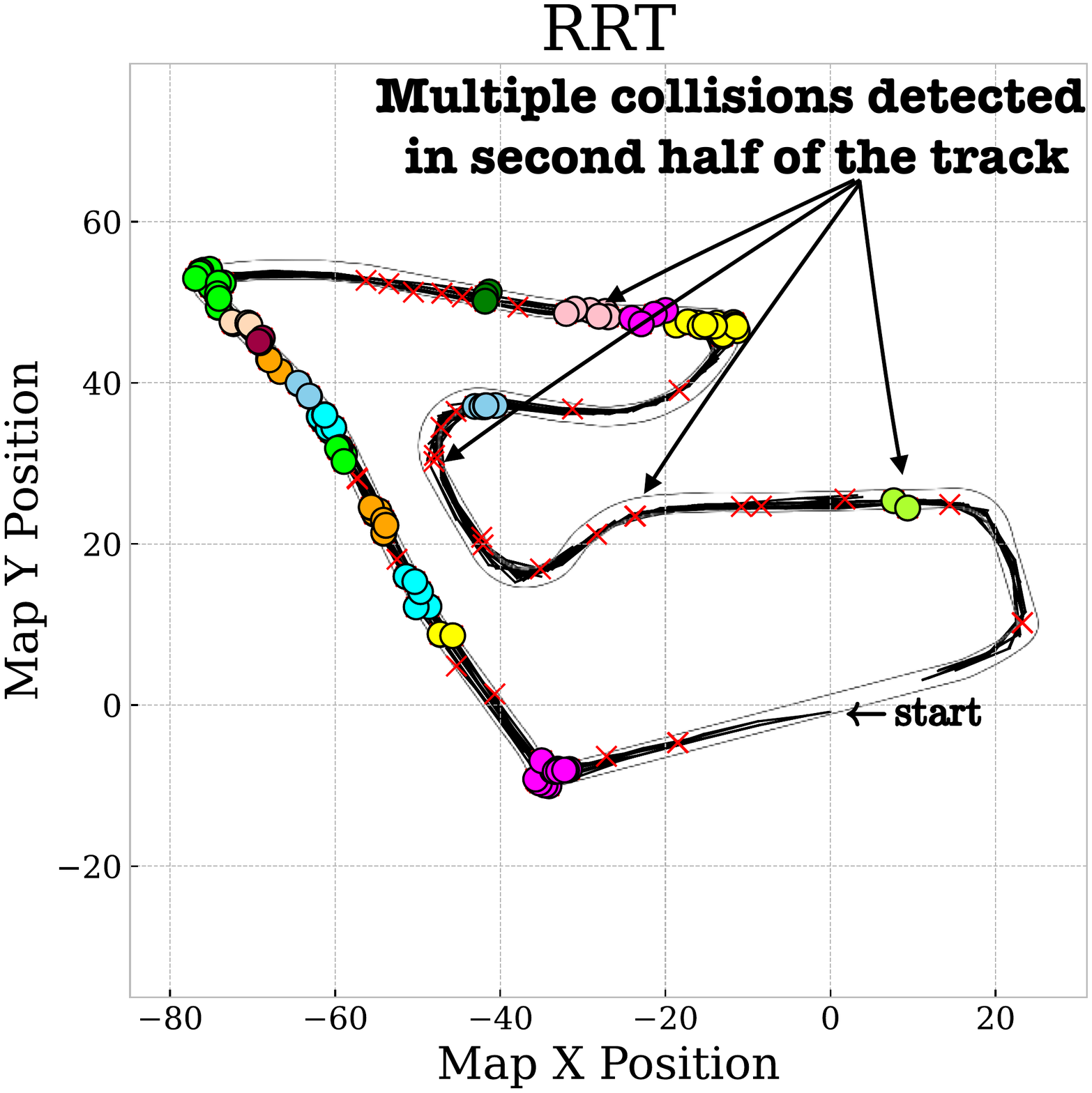}
    \caption{Spatial clustering visualization of the test results for the gap follower planner. Random adversarial perturbations detected 21 unique crashes (8 clusters and 13 outliers), whereas the proposed approach detected 52 unique crashes (17 clusters and 35 outliers). Crash clusters are shown with colored circles and outliers with a red {\color{red} $\times$}.}
    \label{fig:cluster}
    \vspace{-18pt}
\end{figure*}

%In each iteration of the algorithm, a point is sampled randomly in the $n$ dimensional objective space.
%The nearest node on the existing tree to this random point is then chosen using the $L_2$ distance in the objective space.
%Then, the \texttt{expand()} function is called \bz{Need a short description on how the expand function works} to expand the tree and selected a new node to work on.
%Next, we iterate over $\mathcal{V}$ and use each perturbation to generate a short simulation (length $\delta$ in time) which starts from the new node's configuration. These short rollouts are then stored as child nodes of the chosen new node. After the computational budget is exhausted, an $m$-ary tree is created as the output of the algorithm. Finally, the tool provides coverage statistics on the test ran. These statistics helps the end user to find unconventional testing scenarios that wouldn't be discovered by human system designers.

%\bz{this probably belongs in discussion?} With uniform sampling of the search space, the probability of expanding an existing state is proportional to the size of its Voronoi region, hence we would be covering much more unexplored areas and in the view of autonomous racing, we would be discovering much more diverse and unique crashes.
%The \texttt{step\_sim} function is specific to the system under test, thus the stress tester can be generalized over different systems. This in turn simulates the environment starting from a particular point in the objective space for a short time instead of running the full simulation in every test.

\subsection{Autonomy Test Coverage Metrics}
\label{ssec:metrics}
After the process has completed, we compute metrics that provide quantitative feedback about an autonomous algorithm's performance.
These could be used in the future for parameter tuning or comparison of different control strategies.

Specifically, we examine: (1) total failures detected, (2) the spread of the failures in space, recorded by the standard deviation of the failure position, (3) the number of unique failures, measured using the spatial clustering algorithm DBScan~\cite{dbscan}.
The reason to use a clustering algorithm is that we noticed random testing may find a large number of failures, but they are sometimes very similar (for example, crashing on the first turn of a race). 
By clustering results in space, these do not get counted twice, and more information is available on the variety of faults that are found.
Note spatial clustering contains parameters that requires further tuning per application, and the method has two outputs: (i) the number of clusters found and (ii) the number of outliers that do not belong to any cluster.
Adding these two up we compute the number of unique failures.
A preview of our results in the next section with autonomous racing and clustering is shown in Figure~\ref{fig:cluster}, where both random search and the described RRT algorithm have clustered crashes around the turns, but the RRT approach discovers more unique failures.
\section{Experiments}
\label{sec:exp}
% The implementation for the autonomous racing usecase requires implementing what each node will store  and implementing the $step_sim$ function and the $coverage$ function. 
% \begin{enumerate}
%     \item Each node stores a list of children mapped to the command it has come from. The node also stores the positions and the speeds of the vehicles and the status of the simulation if it resulted in a crash or not.
%     \item The $step\_sim$ function takes in the following inputs - 
%     \begin{itemize}
%         \item \textbf{Planner/Controller} - To control the car
%         \item \textbf{Starting\_Position} - Starting position of the two vehicles
%         \item \textbf{Command} - A scalar value which is multiplied with the speed outputed from the Planner/Controller for the opponent vehicle. This ensures variations in the speeds of the ego vehicle which allows the overtaking by the ego vehicle.
%         \item \textbf{Steps} - Number of steps we would be simulating
%     \end{itemize}
%     \item The $coverage$ function will require the only the root node as input. We output the follwoing coverage statistics
%     \begin{itemize}
%         \item Total number of crashes detected
%         \item Maximum Distance Covered by the Ego Vehicle
%         \item Maximum distance between the Opponent and the Ego Vehicle
%         \item Largest Voronoi Area using nodes as f
%     \end{itemize}
% \end{enumerate}

\subsection{Experimental Setup}

We apply the described approach from the previous section to stress test overtake maneuvers in autonomous racing.
Autonomous racing provides a clear objective for the motion planners to drive at high speeds and overtake other agents.
The unstructured operating environment of autonomous racing also makes failure analysis difficult.
%for autonomous path planner we create an experimental environment with an autonomous vehicle in the context of autonomous racing. On the one hand this environment provides a clear objective with driving fast and reliable while overtaking. On the other hand with driving on the racetrack, this repetitive tasks gives us the possibility to compare our approach with others techniques.
For the simulation environment, we use an open source 2D autonomous vehicle simulator created to test path planners and controllers for the F1TENTH autonomous racing competition~\cite{okelly2019}. 
The environment is deterministic with realistic vehicle dynamics modeled by a single track model~\cite{Althoff2017}. 
The physics engine is capable of faster than real-time simulation as well as state serialization (loading and saving), which is needed for our RRT-based algorithm.
In addition, collision with the racetrack boundaries and other vehicles is detected automatically, which we consider as failures.
%We do not consider a 3D environment, since the focus is completely on simple 2D perception and the path planning and control task. Nonetheless
A 2D lidar sensor with noise is also available to enable perception-based algorithms such as object detection and localization methods. 
Finally, the simulation environment is also modular and allows different racetracks and motion planning algorithms to be considered. 

We test the following path planners using the approach:

\begin{itemize}
%	\item \textbf{Race Line Follower:} This algorithms provides a simple path tracker that follows the curvature optimal raceline provided by \cite{Heilmeier2019}.
	\item \textbf{Lane Switcher} creates equispaced lanes that spans the entire track in addition to utilizing a curvature optimal raceline~\cite{Heilmeier2019}. The algorithm makes decision to switch to a specific lane or switch back to the race line when trying to either overtake or block an opponent.
	\item \textbf{Gap Follower} finds gaps in the LiDAR scan by finding the widest range of scan angles that has the highest depth value~\cite{sezer2012novel}. Then it steers the vehicle to follow the largest gap to avoid obstacles.
	\item \textbf{Disparity Extender} is an extension of the Gap Follower where the largest difference in the LiDAR scan is used as the target, with modifications to avoid narrow gaps and walls~\cite{otterness}.
	\item \textbf{Frenet Planner} is based on a semi-reactive method~\cite{Werling2010}. This planner is able to select goal coordinates in the Frenet-Frame of the racetrack and generate multiple trajectories to follow the optimal raceline and avoid obstacles by selecting the appropriate trajectory.
	\item \textbf{Graph Planner} generates a graph covering the race track~\cite{Stahl2019}. The nodes in the graph are vehicle poses in the world frame, and the edges of the graph are generated trajectories similar to those in the Frent Planner. The algorithm selects appropriate actions for the vehicle from the action set for overtaking and following when traversing the graph.
\end{itemize}

Note that for the Disparity Extender, we contacted the University of North Carolina group which won the 2019 F1TENTH race using this algorithm, and were able to use their source code for testing.

\begin{figure*}[t]
    \centering
    \includegraphics[clip, trim=0.4cm 4.8cm 0.1cm 4.9cm, width=0.4\textwidth]{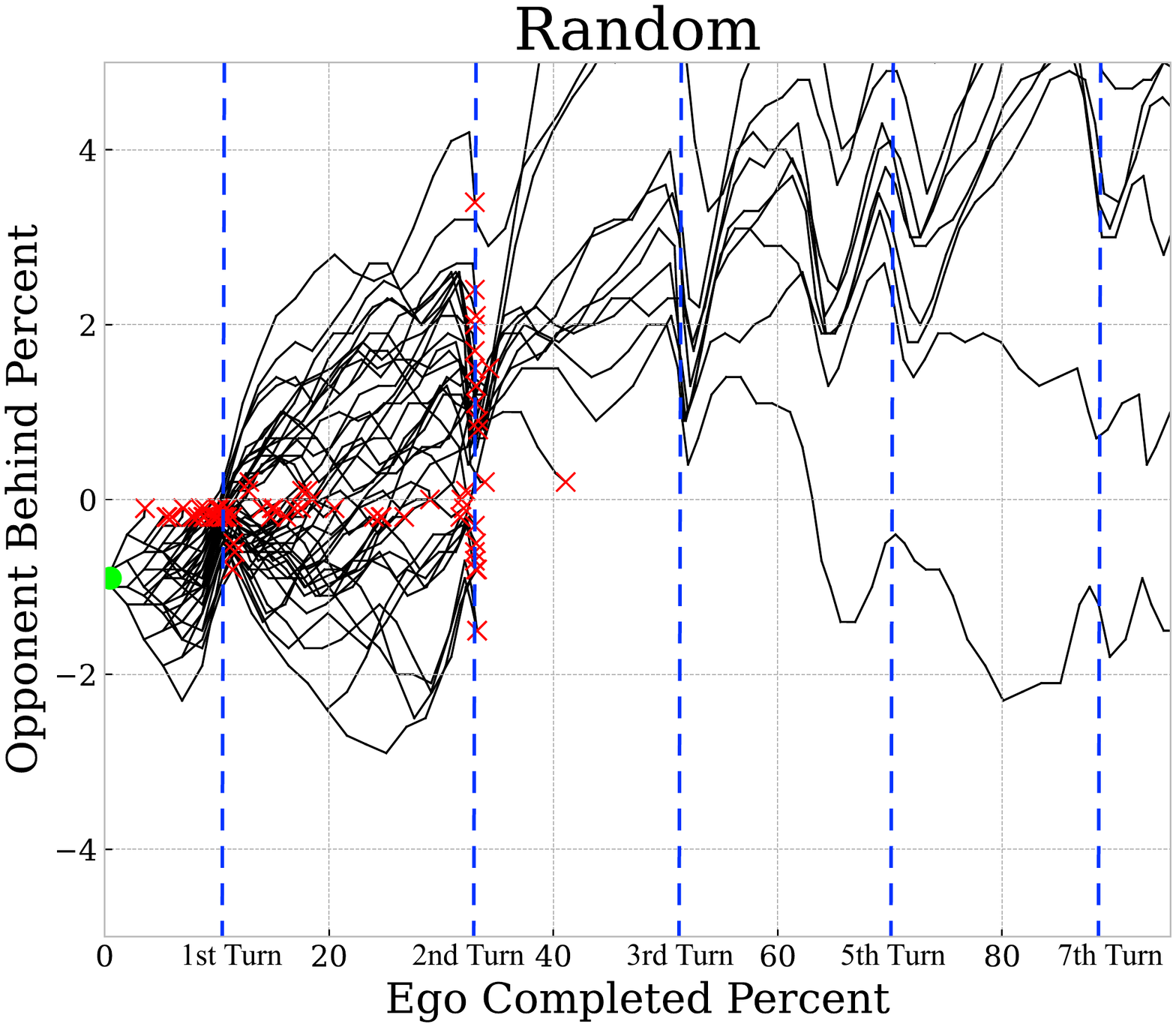} ~~~~~~~~~
    \includegraphics[clip, trim=0.4cm 4.8cm 0.1cm 4.9cm, width=0.4\textwidth]{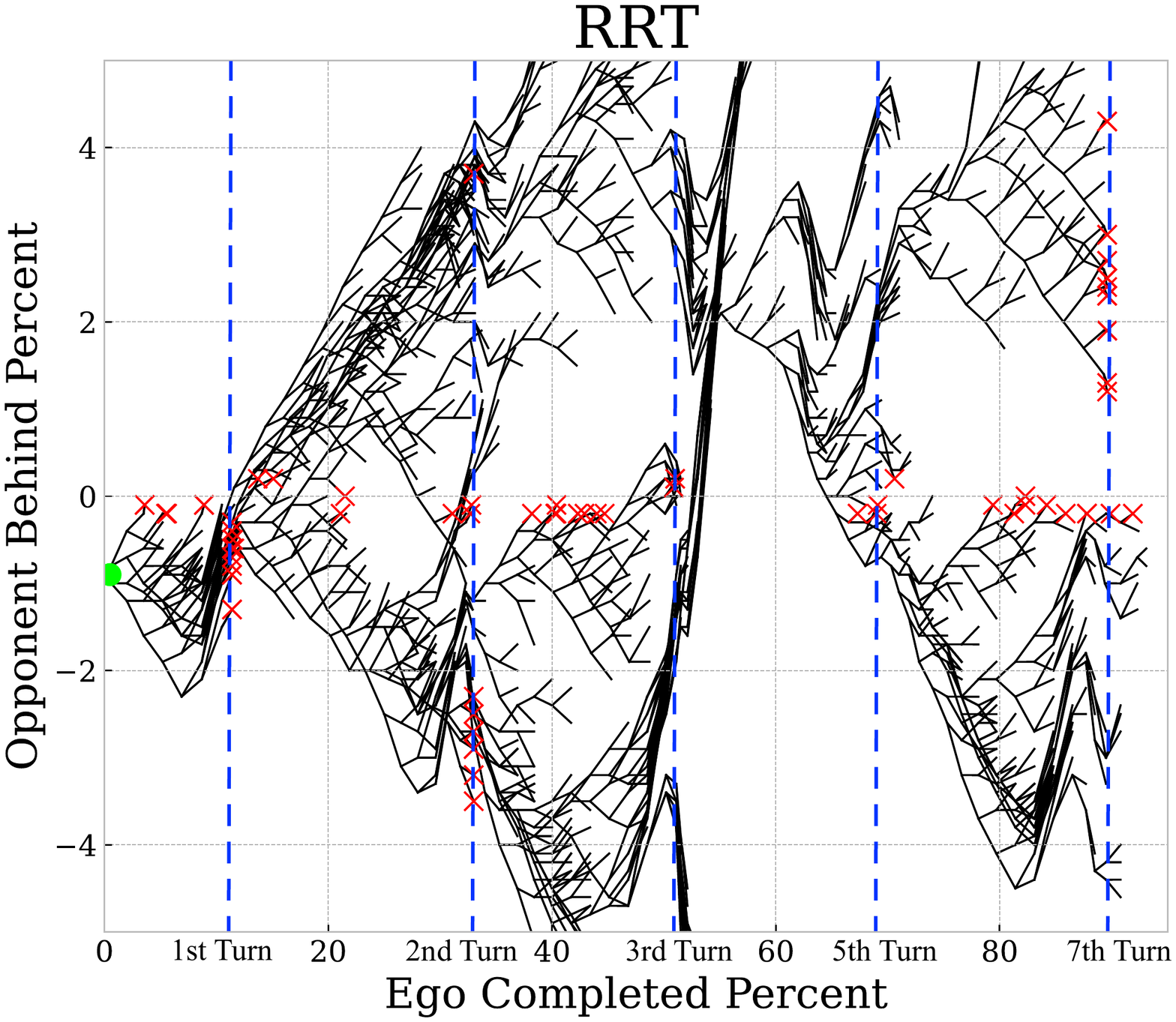}
    \caption{Objective space plots for 2000 seconds of simulation time for the lane switcher planner 
    using random adversarial perturbations (left) and the proposed RRT method (right). 
    All simulations begin at the same initial state  indicated by a green circle at the root of the trees, and collisions are indicated by a red {\color{red} $\times$}.}
    \label{fig:objective_space}
    \vspace{-15pt}
\end{figure*}

To actuate the vehicle to follow created paths we used Pure Pursuit~\cite{coulter1992implementation}, a Stanley controller~\cite{thrun2006stanley} or an LQR controller for path and velocity tracking. 
All testing was done using two vehicles, both of which ran a copy of the same control strategy.
Self-testing is useful in this context as the two cars generally travel a similar speed which leads to more overtake interactions.
Further, if a collision occurs, there is no ambiguity as to which algorithm is at fault.

We next describe the application-specific parameters needed for the general algorithm presented earlier in Section~\ref{sec:generic_stress_tester}.

\vspace{0.5em} \noindent \textbf{Initial Scenario Configuration:} All testing was then done considering a single scenario.
The same track and starting position was used for two vehicles, both of which ran a copy of the same control strategy.
The opponent car started in front of the ego vehicle.
The fixed track start positions define $q_{init}$.

\vspace{0.5em} \noindent \textbf{Adversarial Agent Perturbations:} We perturb the adversarial agent by modifying to the velocity output command of the opponent vehicle.
Two perturbations are considered, $\mathcal{V} = \{opp_{slow}, opp_{fast}\}$, where the former has the opponent moving slightly slower (decreasing the output by 20\%), and the latter has the opponent car moving slightly faster (increasing the output by 20\%).

\vspace{0.5em} \noindent \textbf{Objective Space Projection Function:} We define the objective space as the 2-dimensional space with one axis being the percentage of the race track completed by the ego vehicle, and the other as the percentage of the track that opponent vehicle is ahead of the ego vehicle.
To compute the percentage completion along the track, we take a vehicles $(x, y)$ position and project it onto the center line, and divide by the length of the center line.
The objective space projection function $\mathcal{P}$ does this computation given the positions of the ego and opponent vehicle.

Given the objective space definition, we can see the results of different exploration strategies.
In Figure~\ref{fig:objective_space}, we compare a random search strategy against the proposed RRT approach. 
The random approach has simulations that start at the root and end only after completing a lap or a crash, whereas RRT qualitatively explores the objective space more thoroughly.

\vspace{0.5em} \noindent \textbf{Objective Space Limits:} We use limits $\mathcal{B}$ on the objective space as [-5\%, 5\%] for percentage of opponent vehicle ahead, and [0\%, 95\%] for percentage of ego track completion. 
By limiting the sampling to cases where the cars are close to each other (within 5\% of the track distance), we focus the testing effort to cases where overtake is more likely.

\vspace{0.5em} \noindent \textbf{Simulation Step Function:} The \texttt{step\_sim} function runs 100 steps of the underlying simulation (one second of simulation time), modifying the opponent's velocity command according to the passed-in adversarial agent perturbation.

\begin{table*}[t]
\begin{center}
\caption{Results for five different path planners with both the random and RRT tester. Our RRT test strategy finds more crashes than randomly exploring adversarial agent perturbations.} \label{tab:measurements}
{\setlength{\tabcolsep}{10pt}
\vspace{-5pt}
\begin{tabular}{@{}llllllll@{}}
\toprule
 \textbf{Planner} & \textbf{Tester} & \textbf{\# Crashes} & \textbf{\# 2nd Half Crashes} & \textbf{Pos Stddev}  & \textbf{\# Clusters} & \textbf{\# Outliers} & \textbf{\# Unique Crashes} \\ 
\midrule
  \multirow{2}{1.2cm}{\textbf{Lane Switcher}} 
 & Random & \hspace{0.5em}83.4$\pm$9.6 & \hspace{0.5em}5.2$\pm$0.9 & 13.9$\pm$1.3 & 9.5$\pm$1.9 & 18.4$\pm$4.0 & 27.9 \\  
 & RRT & 254.0$\pm$30.8 (3.0x) & 15.2$\pm$5.8 (2.9x) & 17.9$\pm$1.8 & 9.4$\pm$1.9 & 21.3$\pm$4.6 & 30.7 (1.1x) \\
 \hdashline[0.5pt/3pt]
   \multirow{2}{1.2cm}{\textbf{Gap Follower}} 
    & Random & \hspace{0.5em}65.6$\pm$7.2 & 13.4$\pm$1.8 & 19.2$\pm$1 & 5.4$\pm$0.4 & 21.0$\pm$2.6 & 26.4 \\
   & RRT & 176.8$\pm$19.0 (2.7x) & 84.3$\pm$18.3 (6.3x) & 25.9$\pm$1.2 & \hspace{-0.5em}13.2$\pm$1.4 & 33.7$\pm$4.2 & 46.9 (1.8x) \\
 \hdashline[0.5pt/3pt]
 \multirow{2}{1.2cm}{\textbf{Disparity Extender}} 
 & Random & \hspace{0.5em}73.4$\pm$3.6 & \hspace{0.5em}4.4$\pm$2.1 & 18.2$\pm$1.2 & 4.0$\pm$0.6 & \hspace{0.5em}8.4$\pm$1.3 & 12.4 \\
 & RRT & 342.5$\pm$26.5 (4.7x) & 85.6$\pm$21.0 (19.5x) & 21.3$\pm$1.4 & 8.0$\pm$2.1 & 18.8$\pm$2.9 & 26.8 (2.2x) \\
  \hdashline[0.5pt/3pt]
  \multirow{2}{1.2cm}{\textbf{Frenet Planner}} 
 & Random & 201.4$\pm$21.0 & \hspace{0.5em}2.0$\pm$0.2 & \hspace{0.5em}8.5$\pm$0.9 & 6.3$\pm$2.6 & 17.0$\pm$2.4 & 23.3 \\
  & RRT & 423.3$\pm$10.8 (2.1x) & \hspace{-0.5em}219.1$\pm$6.5 (109.6x) & \hspace{-0.5em}28.39$\pm$0.8 & \hspace{-0.5em}42.3$\pm$2.6 & 27.0$\pm$8.8 & 69.3 (3.0x) \\
 \hdashline[0.5pt/3pt]
  \multirow{2}{1.2cm}{\textbf{Graph Planner}} 
  & Random & \hspace{0.5em}23.8$\pm$4.2 & \hspace{0.5em}1.2$\pm$0.6 & 16.9$\pm$3.2 & 2.0$\pm$0.0 & \hspace{0.5em}2.5$\pm$2.1 & 4.5 \\
  & RRT & \hspace{0.5em}33.5$\pm$3.0 (1.4x) & \hspace{0.5em}3.2$\pm$1.0 (2.6x) & 17.7$\pm$3.9 & 3.5$\pm$0.5 & \hspace{0.5em}2.7$\pm$0.3 & 6.2 (1.4x) \\
 \bottomrule
\end{tabular}
}
\end{center}
%\vspace{-20pt}
\end{table*}

\subsection{Experimental Results}
We next compare the described RRT-based test generation algorithm with a randomized strategy. 
The randomized algorithm starts from the initial position, and randomly consider adversarial agent perturbations at each simulation step.
The method restarts only when either the ego vehicle crashes or a lap is completed. 

We use the autonomy coverage metrics described in Section~\ref{ssec:metrics} to evaluate the methods.
In addition, we add a metric for crashes detected in the second half of the track, as random testing often only got to the second half of the track when the vehicles were far apart and few overtakes occurred.

The spatial clustering DBScan algorithm requires two parameters. For the maximum distance between two samples to be considered in the same cluster we used 2.1 meters.
For the number of samples required to be in the neighborhood to consider it as cluster we used 3.
Visually, these parameters produced results that match our intuition about when crashes our similar, as shown before in Figure~\ref{fig:cluster}.

The importance of excluding similar crashes and not just counting the total number is also apparent in the objective space plot in Figure~\ref{fig:objective_space}.
Here, a large number of crashes are visible at around 10\% and 33\% ego track completion, which correspond to the first two turns on the track (vertical dashed blue lines). For all planners, we performed 10 simulation runs with a stopping condition of 2000 seconds of simulation time, corresponding to 2000 nodes explored for the RRT method.
Each of the 10 experiments used a different random seed and the metrics are then averaged over all the simulations. 

The results are shown in Table~\ref{tab:measurements}.
A result entry of $a \pm b$ shows both the mean $a$ and the standard deviation $b$ for that particular measurement, over all simulation runs. 
The RRT approach was able to find more crashes on all tested planners, up to 4.7x more crashes for the disparity extender. 
In terms of second half crashes, the difference is even more pronounced, with up to 109.6x more crashes being found for the Frenet planner.
More importantly, our method finds more unique crashes, computed as the sum of the mean number of clusters and mean number of outliers found with DBScan, up to 3x more for the Frenet Planner.
An important aspect to keep in mind is that there are diminishing returns with this metric.
Thus, if the RRT method detected twice as many unique crashes, this does not mean that one could just use the random approach for twice as long to get the same result.

It is tempting to compare planning algorithms in the table, as some appear to be safer.
This is not so straightforward, however, as we noticed that planners which crash the least also have the fewest successful overtakes.

In terms of runtime, our approach is fast.
Testing some of the simpler planners like the gap follower or lane switcher up to 2000 seconds of simulation could complete in a few minutes, faster than real-time.
The more complex algorithms, such as the Graph Planner, generally took longer, as the execution of the controller required more computation time.
The bottleneck is typically the execution of the simulation environment and the planner logic, not the test-generation logic, although for planners with significant state the state saving and loading process necessary for RRT can have some impact.
We also could further optimize our implementation through parallelization, as simulations are completely disjoint.

\section{Discussion}
\label{sec:discussion}

One somewhat surprising result of our study was the large number of crashes detected by the different types of planners.
This indicates the difficulty of the problem, and is in line with the outcome of the simulated Indy Autonomous Challenge described in the introduction.

We could likely improve the performance of some of the planners through additional tuning of their parameters.
We did spend some effort on each planner and in single-car races all completed a lap quickly and without crashing.
Manually doing such tuning requires a detailed understanding of each algorithm and is quite burdensome. 
The RRT test-generation method presented here, along with the qualitative autonomous coverage metrics, opens up the possibility of automated parameter tuning of both path planners and controllers in future work, similar to what has been done for single-car planners in other publications~\cite{okelly2020tunercar}.

In terms of our testing strategy, there are several parameters we could change to optimize the search.
Other dimensions could be added to the objective space, such as the horizontal deviation from the center line.
Other adversarial agent perturbations could also be considered such as slight deviations in the angle command or different magnitude changes on the velocity gain.
Rather than testing all perturbations using the simulator, we could also use application-specific logic to select the one that will drive the system to the closest in the objective space, as is more typical with RRT.
Our simulation ran for 100 steps (one second of simulation time) with each call to \texttt{step\_sim}, which could be adjusted to affect the search.
Right now, we also only performed self-testing of a planner against itself.
Other variations could compare planners versus each other.
While these decisions will likely affect the search performance, we did not explore these possibilities in this work. 
We are nonetheless encouraged that using reasonable values produced an efficient test-generation engine.

\section{Conclusion}
\label{sec:conclusion}
In this paper we presented an approach to stress test autonomous path planning algorithms in unstructured high-speed overtake scenarios.
We evaluated five different path planners, include the winner of the 2019 F1TENTH autonomous racing competition.
By perturbing the opponent car's performance slightly, we were able to explore many variations within a fixed scenario, with the goal of finding cases where the ego vehicle crashes.
We demonstrated that randomly searching these variations does detect many crashes, although these are sometimes clustered near the start of the track.
A proposed variant of the RRT path planning algorithm overcomes this issue, increasing number of unique crashes found while keeping the simulation effort constant.
With our approach, the number crashes increased by up to 4.7x, the crashes in the second half of the track increased by up to 109.6x, and the number of unique crashes increased up to 3.0x. 
Our quantitative metrics could be used in the future for automated parameter tuning of path planners to optimize for safety during overtake.

\newpage
\bibliographystyle{IEEEtran}
\bibliography{main,bak,billy}

\end{document}